# SBSS: Stacking-Based Semantic Segmentation Framework for Very High Resolution Remote Sensing Image

Yuanzhi Cai, Lei Fan, and Yuan Fang

*Abstract*—Semantic segmentation of Very High Resolution (VHR) remote sensing images is a fundamental task for many applications. However, large variations in the scales of objects in those VHR images pose a challenge for performing accurate semantic segmentation. Existing semantic segmentation networks are able to analyse an input image at up to four resizing scales, but this may be insufficient given the diversity of object scales. Therefore, Multi Scale (MS) test-time data augmentation is often used in practice to obtain more accurate segmentation results, which makes equal use of the segmentation results obtained at the different resizing scales. However, it was found in this study that different classes of objects had their preferred resizing scale for more accurate semantic segmentation. Based on this behaviour, a Stacking-Based Semantic Segmentation (SBSS) framework is proposed to improve the segmentation results by learning this behaviour, which contains a learnable Error Correction Module (ECM) for segmentation result fusion and an Error Correction Scheme (ECS) for computational complexity control. Two ECS, i.e., ECS-MS and ECS-SS, are proposed and investigated in this study. The Floating-point operations (Flops) required for ECS-MS and ECS-SS are similar to the commonly used MS test and the Single-Scale (SS) test, respectively. Extensive experiments on four datasets (i.e., Cityscapes, UAVid, LoveDA and Potsdam) show that SBSS is an effective and flexible framework. It achieved higher accuracy than MS when using ECS-MS, and similar accuracy as SS with a quarter of the memory footprint when using ECS-SS.

## I. INTRODUCTION

SENMANTIC segmentation is a fundamental task for many remote sensing applications, such as land cover classification, cloud detection and urban scene understanding [1]–[8]. With the rapid development of imaging technology, the resolution of the acquired images has significantly been improved. This trend is also reflected in the publicly available semantic segmentation datasets. For example, the image resolutions are approximately 480p (480×367) in the PASCAL VOC2012 dataset [9], 2k (2048×1024) in the Cityscapes (2016) dataset [10], and 4k (4096×2160 to 3840×2160) in the recently released UAVid (2020) dataset [11]. For a fixed imaging distance, higher camera resolution means finer spatial resolution of an image acquired.

The rich spatial details in Very High Resolution (VHR) images provide an opportunity for more accurate semantic segmentation of a target scene. However, the use of VHR images poses a new challenge to the semantic segmentation task, i.e., the simultaneous segmentation of objects with large scale discrepancies. This is caused by the fact that the fine spatial resolution of VHR images enables segmentation of objects at smaller scales.

Extensive research has been conducted to address this challenge, where deep learning has become the dominant approach. A typical semantic segmentation neural network consists of two components: the encoder and the decoder. The encoder is responsible for extracting features from an input image at multiple down-sampling scales. Subsequently, by interpreting these extracted features, the decoder assigns an appropriate label to each pixel in the image. Various designs have been proposed for these two components to achieve better segmentation performance.

Encoders can broadly be divided into single branch networks and multi-branch networks, according to their macro designs. The representative single-branch encoder networks include VGGNet [12], ResNet [13], Xception [14], MobileNetV2 [15], Swin [16] and ConvNeXt [17], which extract features at progressively reduced scales in a tandem fashion. The representative multi-branch networks include BiSeNet [18] and HRNet [19]. BiSeNet uses the spatial path and the context path to extract high-resolution spatial details and global contextual information, respectively. HRNet uses four parallel branches to extract high-level features at four down-sampling scales simultaneously.

A critical aspect of the decoder design is to expand the receptive field to model long-range dependencies without reducing the spatial resolution. Based on the mechanisms for long-range dependency modelling, decoders can be classified as convolution-based networks and dot-product attention-based networks. The well-known convolution-based ones include U-Net [20], Spatial Pyramid Pooling (SPP) [21], Pyramid Pooling Module (PPM) [22] and Atrous Spatial Pyramid Pooling (ASPP) [23]–[25], which rely mainly on the pyramid-like structure. Meanwhile, the success of dot-product attention-based networks is due to their global modelling ability.

This research was funded by Xi'an Jiaotong-Liverpool University Research Enhancement Fund, grant number REF-21-01-003, Xi'an Jiaotong-Liverpool University Key Program Special Fund, grant number KSF-E-40, and Xi'an Jiaotong-Liverpool University Research Development Fund, grant number RDF-18-01-40. *(Corresponding author: Lei Fan).*

Yuanzhi Cai, Lei Fan, and Yuan Fang are with Department of Civil Engineering, Design School, Xi'an Jiaotong-Liverpool University, Suzhou, 215000, China (e-mail: yuanzhi.cai19@student.xjtlu.edu.cn; lei.fan@xjtlu.edu.cn; yuan.fang16@student.xjtlu.edu.cn).

However, the computational cost of the dot-product attention mechanism increases quadratically with the size of the features used [26]. Therefore, an efficient use of this mechanism is a main focus of the relevant previous research, which includes Dual Attention Network (DANet) [27], Disentangled Non-Local Neural Networks (DNLNet) [28], Object Context network (OCRNet) [29], Attentive Bilateral Contextual network (ABCNet) [27] and Multiattention Network (MANet) [30].

Although there are various designs of segmentation networks, they share one common characteristic, which is that features can only be extracted on a predefined set of scales. Due to computational constraints, it is unpractical to extract features at too many scales (currently up to four scales) in a segmentation network. However, there is no guarantee that those pre-defined scales are the optimal ones for a given application scenario. To enable a segmentation network to analyse images at a wider range of scales, it is the common practice to use a test-time data augmentation called the Multi Scale (MS) test. The MS resizes the original images to various scales and feeds them into a segmentation network. The output segmentation maps are then often assembled by average voting. In addition, the MS is typically used in conjunction with training-time multi-scale data augmentation. When both methods are used, the entire segmentation framework falls under an ensemble learning technique called bootstrap aggregating (Bagging) [31]. More specifically, images at different scales are used as the bagging samples in this process. The role of bagging is to reduce the variance of errors among multiple predictions using different bagging samples [32], [33]. In other words, the MS works best if the prediction error for each class is randomly distributed over the scales used for the image resizing. The MS has become the default method for testing the best performance of a segmentation network. However, it is worth investigating whether there is a better method to fuse the segmented maps resulting from input images of different resizing scales.

It is common sense that the size distribution of objects of different classes in an image is similar to that in reality, despite the effects of perspective. For example, larger objects in reality are usually also larger in images (e.g., buildings often occupy a large area in street view images). Meanwhile, studies have shown that the size of the effective receptive field for a network is limited [34]–[36]. Therefore, it is hypothesised that for a given segmentation network and a dataset, each class may have its preferred resizing scale for segmentation. More specifically, the classes that typically have large objects may prefer to be shrunk so that they can be fitted into the effective receptive field and segmented as a whole, while the classes that typically have small objects may prefer to be zoomed in to avoid becoming indistinguishable after being downsampled by the segmentation network. In other words, the prediction errors for each class may have biases related to the resizing scales.

Based on this hypothesis and inspired by previous studies [37], [38], a Stacking-Based Semantic Segmentation (SBSS) framework was proposed in this study to reduce the error associated with the resizing scales. In the SBSS framework proposed, a segmentation map obtained at the smaller scale is gradually corrected by a learnable Error Correction Module (ECM) using a segmentation map obtained at a larger resizing scale. This process starts with an initial segmentation map at the smallest scale considered and is repeated multiple times (each time, a larger scale was used). The computational complexity of the SBSS framework is flexible, which can be altered by assigning different Error Correction Schemes (ECS). In particular, two ECS were proposed in this study, namely ECS-MS and ECS-SS, which have similar Floating-point operations (Flops) to the MS test and the Single-Scale (SS) test, respectively. The ECS-MS is designed for the applications requiring the highest possible segmentation accuracy. Meanwhile ECS-SS is designed for the applications where Graphics Processing Unit (GPU) memory is limited. The effectiveness of the SBSS framework was demonstrated on four datasets, including Cityscapes, UAVid, LoveDA and Potsdam, which cover a variety of scenarios (e.g., urban and rural) and acquisition perspectives (e.g., street levels, inclined drones and aerial views).

## II. SBSS FRAMEWORK

### A. Overview of SBSS framework

For $n$ selected scales (sorted from the smallest to the largest), the workflow of the SBSS framework is illustrated in Fig. 1, where $i = 1 \sim n$. The explanations of the abbreviations used are listed in Table I. To implement SBSS, an initial segmentation map ($Y_1$) is required, which is obtained using the following two steps: (1) an original input image is resized to the smallest scale considered; (2) the resized input image ($X_1$) is fed to a segmentation network to obtain the initial segmentation map.

The segmentation map ($Y_i$) at the beginning of the workflow is used in two parallel processes. In one process, $Y_i$ is simply resized to a next scale to obtain the resized map $Y_{i \rightarrow i+1}$. In the other process, $Y_i$ is fed into an error correction scheme (detailed in Section II-C) to determine the area(s) where error correction is required (for ease of demonstration, only one local area is shown in Fig. 1). Once the local area is identified, it is used to crop the local segmentation information $Y_{i,s}$ from $Y_i$, and meanwhile to crop the local image $X_{i+1,s}$ from the input image ($X_{i+1}$) that is resized to the $i+1$ scale. The resized input image of the local area is also fed into the segmentation network to obtain the its corresponding segmentation information $Y_{i+1,s}$. The two segmentation maps (i.e., $Y_{i,s}$ and $Y_{i+1,s}$) of the selected area are processed in an error correction module (detailed in Section II-B), which result in a corrected map $Y_{i+1,s,c}$ of the selected area. The segmentation information in $Y_{i+1,s,c}$ is used to replace that in the corresponding pixels in $Y_{i \rightarrow i+1}$, which produces an updated segmentation map $Y_{i \rightarrow i+1,s,c}$. If additional rounds of the iteration process are considered, $Y_{i \rightarrow i+1,s,c}$ is essentially the segmentation map ($Y_i$) used at the beginning of the workflow in the next round. Otherwise, it is the output (i.e., $Y_{n-1 \to n,s,c}$) of the last round. The final segmentation map is obtained by resizing $Y_{n-1 \to n,s,c}$ to the size of the original input image.

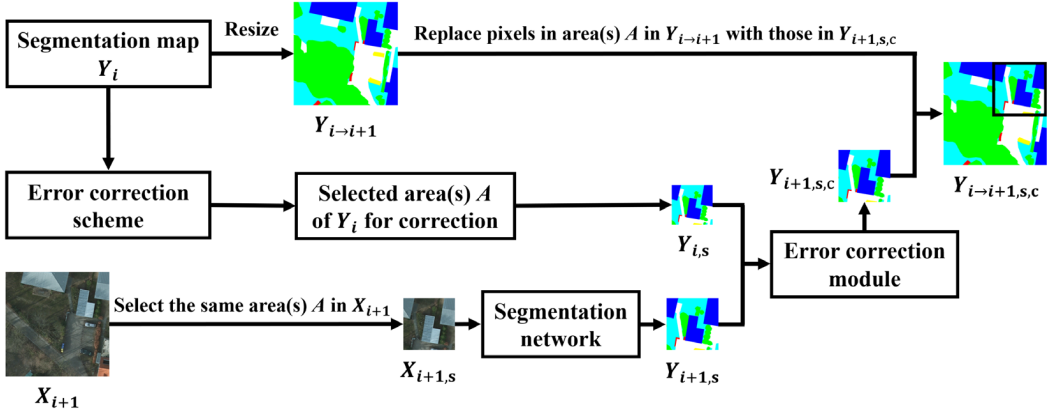

**Fig. 1.** Workflow of one iteration of the SBSS framework, in which $Y_i$ and $X_{i+1}$ are the input maps; $Y_{i \to i+1,s,c}$ is the output map of one iteration and the segmentation map (i.e., a new $Y_i$) for the next iteration; the loop is ended when $i + 1 = n$ (i.e., the number of scales used in SBSS); the area(s) $A$ are determined by the error correction scheme, and are also applied to $X_{i+1}$ and $Y_{i \to i+1}$.

TABLE I
THE EXPLANATIONS OF THE ABBREVIATIONS USED IN FIG. 1

| Abbreviations | Explanations |
|---|---|
| $Y_i$ | The segmentation map of the $i$th scale. |
| $Y_{i \to i+1}$ | The segmentation map resized from the $i$th to the $(i+1)$th scale. |
| $Y_{i,s}$ | The selected area in $Y_i$ for error correction. |
| $X_{i+1}$ | The input image that is resized to the $(i+1)$th scale. |
| $X_{i+1,s}$ | The selected area from $X_{i+1}$ for error correction. |
| $Y_{i+1,s}$ | The segmentation map obtained using $X_{i+1,s}$ as the input. |
| $Y_{i+1,s,c}$ | The corrected map $Y_{i+1,s}$ of the selected area. |
| $Y_{i \to i+1,s,c}$ | The corrected map $Y_{i \to i+1}$. |

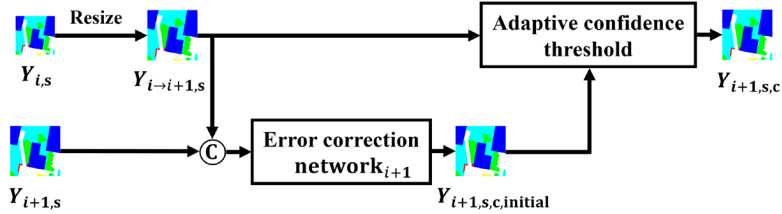

**Fig. 2.** Error correction module.

There are three main components in the SBSS framework, including a segmentation network, ECS and ECM. The segmentation network can be any existing network that provides a reasonably good initial segmentation result. The ECS is also flexible. For example, one can choose to correct the entire segmentation map or only selected areas within it. All of these allow SBSS to be applied to various application scenarios with different demands. More detailed descriptions of the ECM and the two proposed ECS (ECS-MS and ECS-SS) are given in Sections II-B and II-C respectively.

### B. Error correction module

The proposed error correction module is shown in Fig. 2. The segmentation map ($Y_{i,s}$) at a lower scale is first resized to a higher scale to obtain the resized map $Y_{i \to i+1,s}$. $Y_{i \to i+1,s}$ is concatenated with the segmentation map $Y_{i+1,s}$ using the input image at a higher scale. Subsequently, they (i.e., $Y_{i+1,s}$ and

$Y_{i \to i+1,s}$) are fed into an Error Correction Network (ECN) to obtain the initial corrected segmentation map ($Y_{i+1,s,c,initial}$). Finally, an Adaptive Confidence Threshold (ACT) is used to replace the corresponding pixels in $Y_{i \to i+1,s}$ with the more confident pixels in $Y_{i+1,s,c,initial}$ to obtain corrected segmentation information $Y_{i+1,s,c}$.

#### 1) Error correction network

The detailed structure of the ECN is presented in Fig. 3. For a dataset having $C$ classes, its segmentation map also has $C$ channels. Each channel of the segmentation map records the segmentation probability of its corresponding class. Therefore, concatenating the two segmentation maps will result in a feature map with channel number of $2C$, which is used as the input to the ECN. The input features are processed through the stem block and two residual blocks. The resulting feature map (96 channels) is then compressed by a pointwise convolution layer (with $C$ kernels) to output the initial corrected segmentation map ($Y_{i+1,s,c,initial}$). The weights of the ECN are not shared across scales (i.e., the ECN is trained separately for each scale).

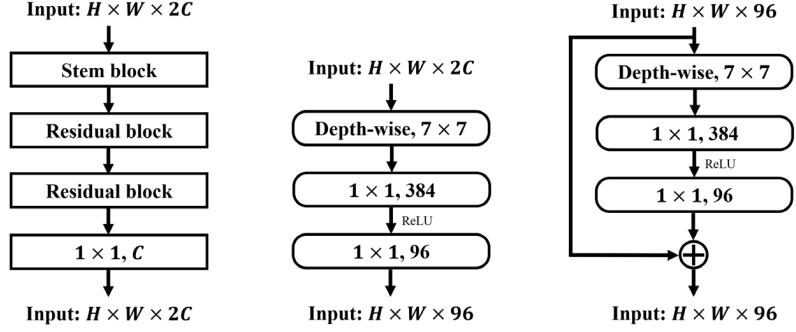

**Fig. 3.** Structures of the error correction network (left), the stem block (middle), and the residual block (right).

TABLE II

COMPARISON OF THE TOTAL SIZE OF THE IMAGES TO BE PROCESSED BY MS AND SBSS-MS

| | Scales | 0.5 | 0.75 | 1.0 | 1.25 | 1.5 | 1.75 | Sum |
|---|---|---|---|---|---|---|---|---|
| MS | Selection percentage of non-overlapping patches | 100% | 100% | 100% | 100% | 100% | 100% | - |
| | Ratio of the total size of the patches to the original image | 25% | 56% | 100% | 156% | 225% | 306% | 869% |
| ECS-MS | Selection percentage of non-overlapping patches | 100% | 100% | 100% | 100% | 100% | 0% | - |
| | Ratio of the total size of the patches to the original image | 25% | 56% | 100% | 156% | 225% | 0% | 563% |

The structure of the residual block is designed in reference to the basic block of ConvNeXt [17]. Similar to ConvNeXt's study we found that better results were achieved using blocks with large convolutional kernels than those with small 3x3 kernels (e.g., ResNet block). The other minor modification is the removal of the normalisation layer. This is because a small decrease in segmentation accuracy was observed when using the normalisation layer in our study.

### 2) Adaptive confidence threshold

The objective of using the ACT is to select pixels that have low confidence levels in $Y_{i \to i+1,s}$ but high confidence levels in the initial corrected segmentation map ($Y_{i+1,s,\text{initial}}$). The ACT is implemented as follows. For each pixel in a segmentation map, its value in each channel represents the confidence level of belonging to the corresponding class, and the values in all channels are summed to one. Thus, the confidence level of each pixel in $Y_{i+1,s,c,\text{initial}}$ is its maximum value over all channels. The confidence map for $Y_{i+1,s,c,\text{initial}}$ and $Y_{i \to i+1,s}$ are denoted as $Y_{i+1,s,c,\text{initial}}^{\text{confidence}}$ and $Y_{i \to i+1,s}^{\text{confidence}}$, respectively. An Adaptive Confidence ($AD$) map is produced to match the objective of the adaptive confidence threshold, as shown in Equation (1) where ".∗" represents the element-wise multiplication.

$$AD = \left(1 - Y_{i \to i+1,s}^{\text{confidence}}\right).* Y_{i+1,s,c,\text{initial}}^{\text{confidence}} \qquad (1)$$

The ACT was set to be the median pixel value in the $AD$ map. Finally, the regions in the $AD$ map that exceed the threshold are recorded and the results within that region in $Y_{i \to i+1,s}$ are replaced with the results in $Y_{i+1,s,c,\text{initial}}$ to obtain $Y_{i+1,s,c}$.

### C. Error correction scheme

In the SBSS framework, apart from the segmentation network used, ECS also has a significant impact on the overall computational load. The commonly used metrics for quantifying computational load include Flops and GPU memory footprints. For a given network structure, the main factors affecting Flops and GPU memory footprints are the total number and the size of the input patches, respectively. Therefore, the focus of developing an ECS is on the selection of areas to be corrected and on the choice of a patch size that does not exceed the GPU memory limit. With reference to the Flops required for the two commonly used test methods (i.e., MS and SS), two ECS (ECS-MS and ECS-SS) are proposed in this study. It is worth noting that the SBSS framework using ECS-MS and ECS-SS are abbreviated as SBSS-MS and SBSS-SS in subsequent sections. Moreover, for image patches extraction, the non-overlapping sliding window approach is used in this study.

### 1) Error correction scheme with Flops at the multi-scale test level

The MS has widely been adopted to obtain the highest possible segmentation accuracy. The commonly used set of scales is {0.5, 0.75, 1.0, 1.25, 1.5, 1.75}[19], [39]. Under this setting, the original image is resized using each of those six scales before being processed by a segmentation network. The total size of the image patches that need to be processed by MS is shown in Table II, which is almost nine times that of the original image. The proposed ECS-MS also processes all the image patches at each scale level, but with less scales. As shown in Table II, the scale 1.75 is discarded in ECS-MS, which is to compensate for the additional Flops for ECM. Since the Flops of ECM are quite small compared to that of the segmentation network, the current ECS-MS setup is conservative. The exact Flops for ECS-MS and MS are given in Section III-E.

### 2) Error correction scheme with Flops at the single-scale test level

The SS is often used when computational resources are limited, which only analyses the original image (i.e., at scale 1.0). Based on such considerations, the ECS-SS is designed in this study. The ECS-SS analyses images at four scales:

TABLE III

COMPARISON OF THE TOTAL SIZE OF THE IMAGES TO BE PROCESSED BY SS AND SBSS-SS

| | Scales | 0.25 | 0.5 | 1.0 | 1.5 | Sum |
|---|---|---|---|---|---|---|
| SS | Selection percentage of non-overlapping patches | 0% | 0% | 100% | 0% | - |
| | Ratio of the total size of the patches to the original image | 0% | 0% | 100% | 0% | 100% |
| ECS-SS | Selection percentage of non-overlapping patches | 100% | 100% | 25% | 8.33% | - |
| | Ratio of the total size of the patches to the original image | 6.25% | 25% | 25% | 18.75% | 75% |

TABLE IV

SUMMARY OF FOUR DATASETS USED

| Dataset | Number of images | | | Number of classes | Image resolution |
|---|---|---|---|---|---|
| | Training set | Validation set | Test set | | |
| Cityscapes | 2975 | 500 | 1525 | 19 | 2048×1024 |
| UAVid | 200 | 70 | 150 | 8 | 4096×2160 or 3840×2160 |
| LoveDA | 2522 | 1669 | 1796 | 7 | 1024×1024 |
| Potsdam | 24 | - | 14 | 6 | 6000×6000 |

TABLE V

TRAINING SETTING FOR THE SEGMENTATION NETWORK

| Dataset | Cityscapes | UAVid | LoveDA | Potsdam |
|---|---|---|---|---|
| Patch size | 1024×512 | 1024×540 | 512×512 | 750×750 |
| Total training iterations | 80 k | 80 k | 15 k | 15 k |
| Pretraining dataset | ImageNet-1k | | | |
| Optimizer | Stochastic Gradient Descent (SGD) | | | |
| Initial learning rate | 0.01 | | | |
| Learning rate schedule | Poly learning rate policy with a power of 0.9 | | | |
| Momentum | 0.9 | | | |
| Weight decay | 0.0005 | | | |
| Batch size | 16 | | | |
| Loss function | Cross entropy | | | |
| Data augmentation | Random cropping, random resize (0.25~2), random horizontal flipping, photo metric distortion | | | |

{0.25, 0.5, 1.0, 1.5}. To keep the flops consumed by ECS-SS similar to SS, ECS-SS is unable to analyse all the image patches at four scales. Therefore, a selection strategy is designed for ECS-SS to analyse only part of the image patches. As shown in Table III, for the two smaller scales (i.e., 0.25 and 0.5), all image patches are selected because of their relatively small total size compared to the original image. While for the latter two scales (i.e., 1.0 and 1.5), only part of the image patches is selected. The selection is based on the confidence map at that scale (i.e., $Y_{i \to i+1,s}^{confidence}$). Patches with relatively low confidence accumulations are selected for analysis. With this setup, SBSS-SS allows the use of images at a wider range of scales for analysis while keeping the total size of the images to be processed at 75% of that of SS. Similar to ECS-MS, the Flops saved by using ECS-SS are a compensation for the extra Flops involved in ECM, and the exact Flops are provided in Section III-E.

## III. EXPERIMENTS AND RESULTS

### A. Datasets and implementation details

#### 1) Datasets

The effectiveness of the proposed SBSS framework was tested on four datasets, including Cityscapes, UAVid, LoveDA and Potsdam. The key characteristics of these datasets are summarised in Table IV. The partition of the training, validation and test sets for the first three datasets follows their original implementations. For the Potsdam dataset, the RGB images with IDs of 2_13, 2_14, 3_13, 3_14, 4_13, 4_14, 4_15,

5_13, 5_14, 5_15, 6_13, 6_14, 6_15 and 7_13 were used as the testing set (the same set of images was also used as the validation set in this study), while the remaining 24 RGB images were used for training.

#### 2) Training setting

The training settings used in this study are summarized in Table V. Most of these settings were consistent across three datasets used. The crop size used was different as it was set to be proportional to the original image in the dataset. For the UAVid dataset which has two different image sizes, all images were resized to 4096 × 2160 for ease of processing. The total training iterations for LoveDA and Potsdam were significantly less than those of the other two datasets, due to the relatively small sizes of LoveDA and Potsdam.

After the segmentation networks had been trained, they were used to generate segmentation maps for each scale required to train the ECN. The ECN was trained using settings similar to those in Table V. The differences include: no pretraining, no random scaling and photometric distortion being used, the total number of training iterations, and the initial learning rate that was reduced to one tenth of that in Table V.

#### 3) Evaluation metrics

The segmentation accuracy was evaluated using mean Intersection over Union (mIoU) in this study. Based on the confusion matrix, the mIoU is computed as:

$$mIoU = \frac{1}{C} \sum_{c=1}^{C} \frac{TP_c}{TP_c + FP_c + FN_c} \quad (2)$$

Where $TP_c$, $FP_c$, and $FN_c$ represent the true positive, false

TABLE VI
Input scales that achieve the highest segmentation accuracy for different classes

| Method | Backbone | Cityscapes (Patch size of 1024×512) | | | | UAVid (Patch size of 1024×540) | | | |
|---|---|---|---|---|---|---|---|---|---|
| | | Road | Sidewalk | Person | Bicycle | Building | Tree | Static car | Human |
| FCN | HRNet-w18 | 0.75 | 0.75 | 1.50 | 1.50 | 0.75 | 0.50 | 0.75 | 1.75 |
| BiSeNetV1 | ResNet50 | 0.75 | 0.75 | 1.75 | 1.50 | 0.75 | 0.50 | 1.00 | 1.75 |
| PSPNet | ResNet50 | 0.75 | 0.75 | 1.50 | 1.50 | 0.50 | 0.50 | 1.00 | 1.75 |
| DeepLabV3+ | ResNet50 | 0.75 | 0.75 | 1.50 | 1.25 | 0.50 | 0.50 | 0.50 | 1.50 |
| DANet | ResNet50 | 0.75 | 0.75 | 1.75 | 1.50 | 0.50 | 0.75 | 1.00 | 1.50 |
| GCNet | ResNet50 | 0.75 | 0.75 | 1.50 | 1.50 | 0.75 | 0.50 | 1.00 | 1.50 |
| DNLNet | ResNet50 | 0.75 | 0.75 | 1.50 | 1.50 | 0.75 | 0.75 | 1.00 | 1.25 |
| UperNet | ResNet50 | 0.75 | 0.75 | 1.50 | 1.50 | 0.75 | 0.50 | 1.00 | 1.75 |
| UperNet | Swin-T | 0.75 | 0.75 | 1.75 | 1.50 | 0.50 | 0.75 | 1.00 | 1.75 |
| UperNet | ConvNeXt-T | 0.75 | 0.75 | 1.75 | 1.50 | 0.75 | 0.50 | 0.75 | 1.25 |
| Method | Backbone | LoveDA (Patch size of 512×512) | | | | Potsdam (Patch size of 750×750) | | | |
| | | Agricultural | Water | Forest | Barren | Building | Impervious | Tree | Car |
| FCN | HRNet-w18 | 0.50 | 0.50 | 1.00 | 1.00 | 0.75 | 1.00 | 1.00 | 1.50 |
| BiSeNetV1 | ResNet50 | 0.75 | 0.75 | 1.25 | 1.00 | 1.00 | 1.00 | 1.25 | 1.50 |
| PSPNet | ResNet50 | 0.50 | 0.75 | 1.25 | 1.00 | 0.75 | 1.00 | 1.25 | 1.50 |
| DeepLabV3+ | ResNet50 | 0.50 | 0.50 | 1.25 | 1.00 | 0.75 | 1.00 | 1.25 | 1.25 |
| DANet | ResNet50 | 0.50 | 0.50 | 1.50 | 1.00 | 0.75 | 1.00 | 1.00 | 1.25 |
| GCNet | ResNet50 | 0.50 | 0.75 | 1.25 | 1.00 | 0.75 | 1.00 | 1.00 | 1.25 |
| DNLNet | ResNet50 | 0.50 | 0.75 | 1.25 | 1.00 | 0.75 | 1.00 | 1.25 | 1.25 |
| UperNet | ResNet50 | 0.50 | 0.75 | 1.25 | 1.00 | 0.75 | 1.00 | 1.00 | 1.50 |
| UperNet | Swin-T | 0.50 | 0.75 | 1.50 | 1.00 | 0.75 | 1.25 | 1.00 | 1.25 |
| UperNet | ConvNeXt-T | 0.50 | 0.75 | 1.75 | 1.00 | 0.75 | 1.00 | 1.25 | 1.25 |

TABLE VII
Segmentation accuracy (mIoU) on validation sets using single scale tests (%)

| Method | Backbone | Cityscapes | | UAVid | | LoveDA | | Potsdam | |
| | | Patch size | | Patch size | | Patch size | | Patch size | |
| | | 1024×512 | 512×256 | 1024×540 | 512×270 | 512×512 | 256×256 | 750×750 | 375×375 |
|---|---|---|---|---|---|---|---|---|---|
| FCN | HRNet-w18 | 75.73 | 68.85 | 73.75 | 72.40 | 51.20 | 49.60 | 85.49 | 76.69 |
| BiSeNetV1 | ResNet50 | 75.06 | 58.55 | 73.19 | 71.60 | 49.36 | 45.90 | 84.70 | 81.78 |
| PSPNet | ResNet50 | 77.90 | 72.68 | 73.42 | 72.03 | 51.49 | 49.74 | 85.85 | 84.26 |
| DeepLabV3+ | ResNet50 | 78.66 | 74.35 | 73.65 | 72.40 | 50.71 | 48.72 | 85.73 | 84.22 |
| DANet | ResNet50 | 78.64 | 74.36 | 73.63 | 72.54 | 51.36 | 50.28 | 86.06 | 84.98 |
| GCNet | ResNet50 | 77.68 | 73.49 | 73.33 | 72.22 | 50.80 | 49.64 | 85.82 | 84.70 |
| DNLNet | ResNet50 | 78.31 | 74.50 | 73.47 | 72.33 | 51.25 | 50.27 | 85.61 | 84.65 |
| UperNet | ResNet50 | 77.65 | 71.92 | 73.94 | 72.34 | 51.04 | 48.75 | 85.62 | 83.62 |
| UperNet | Swin-T | 77.47 | 74.99 | 74.06 | 72.57 | 52.42 | 50.30 | 86.07 | 85.01 |
| UperNet | ConvNeXt-T | 78.84 | 75.52 | 74.26 | 72.68 | 52.52 | 50.47 | 86.41 | 85.12 |

positive and false negatives of class $c$, respectively.

### B. Scale related segmentation error

The study is based on the hypothesis that different classes have their preferences for the resizing scale used. Extensive experiments were conducted in this study to confirm the validity of this hypothesis, which are presented in this section.

In total, ten segmentation networks of similar sizes were tested in this study, including HRNet [19], BiSeNetV1 [18], PSPNet [22], DeepLabV3+ [25], DANet [27], GCNet [40], DNLNet [28], UperNet [41], Swin [16], and ConvNeXt [17]. These networks are representative works in the field of semantic segmentation. In the experiments, these networks were trained on the training sets of the four datasets. For each dataset, the input images in the validation set were resized to a set of scales {0.5, 0.75, 1.0, 1.25, 1.5, 1.75}, generating six new validation sets at these scales. The SS tests were performed on these newly generated validation sets. The scale corresponding to the highest segmentation accuracy obtained for each class was recorded. To facilitate the presentation of the experimental results in reasonably sized tables, we have randomly selected four classes in each dataset and presented their preferred resizing scales in Tables VI. It was observed that the preferred resizing scale for a class is usually the opposite of the size of the image area occupied by that class. For example, the classes that usually occupy larger areas in the image prefer to be segmented using smaller resizing scales. These experimental results proved the validity of the hypothesis of this study.

### C. Segmentation network choice

In the SBSS framework, the role of the segmentation network is to provide the raw segmentation map at multiple scales. A more accurate raw segmentation map is beneficial to improve the final segmentation accuracy of the SBSS framework. The segmentation accuracies of those ten networks (presented in Section III-B) using SS on the validation set is summarised in Table VII. The highest segmentation accuracies were achieved by ConvNeXt-T in all tests, and was therefore chosen as the segmentation network for the SBSS framework in this study.

TABLE VIII
THE QUANTITATIVE RESULTS OF THE ABLATION STUDIES ON VALIDATION SETS OF FOUR DATASETS

| Method | Cityscapes | | UAVid | | LoveDA | | Potsdam | |
|---|---|---|---|---|---|---|---|---|
| | Patch size | mIoU (%) | Patch size | mIoU (%) | Patch size | mIoU (%) | Patch size | mIoU (%) |
| MS | | 81.32 | | 75.76 | | 53.18 | | 87.34 |
| ECS-MS + ACT | 1024×512 | 81.64 | 1024×540 | 76.14 | 512×512 | 53.91 | 750×750 | 87.43 |
| ECS-MS + ECN | | 82.82 | | 76.53 | | 55.77 | | 87.61 |
| SBSS-MS (ECS-MS + ACT + ECN) | | 83.05 | | 76.78 | | 56.28 | | 87.68 |
| SS | | 75.52 | | 72.34 | | 50.47 | | 85.12 |
| ECS-SS + ACT | 512×256 | 76.15 | 512×270 | 72.73 | 256×256 | 51.03 | 375×375 | 85.45 |
| ECS-SS + ECN | | 78.11 | | 73.57 | | 51.66 | | 86.15 |
| SBSS-SS (ECS-SS + ACT + ECN) | | 78.52 | | 73.85 | | 51.98 | | 86.35 |

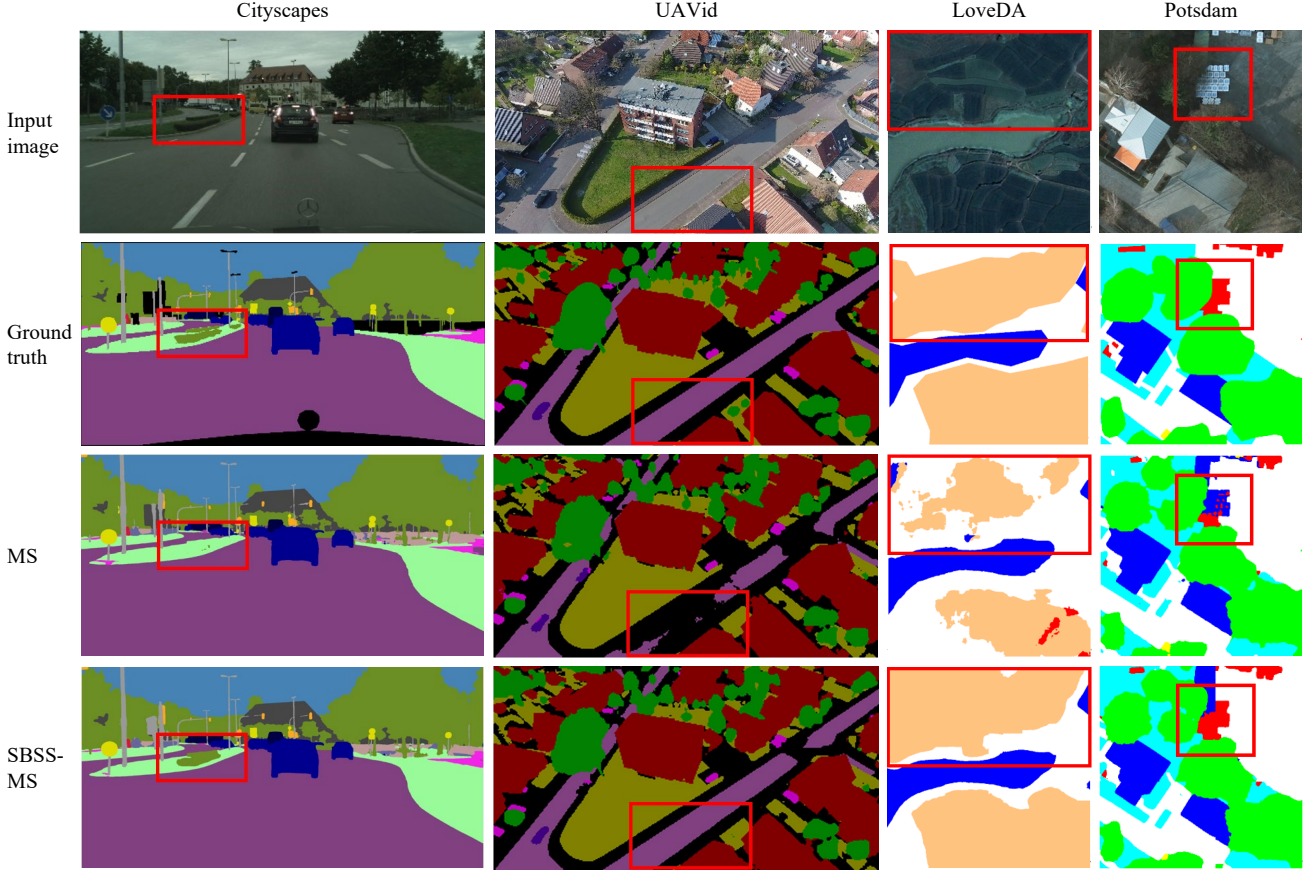

**Fig. 4.** Qualitative comparisons between MS and SBSS-MS on the Cityscapes, UAVid, LoveDA and Potsdam validation sets.

### D. Ablation study on the test setting

To evaluate the effectiveness of the different components within the SBSS framework, extensive ablation experiments were conducted in this study. The experimental setup and the corresponding results are presented in Table VIII. For each dataset, experiments were conducted with two input patch sizes. The larger one represented the input size typically used in previous studies. The other input size was half of the larger one, which was used to simulate the scenario of limited GPU memories.

As shown in Table VIII, using either the ACT or the ECN alone improved the segmentation accuracy by an average of 0.43% or 1.40% respectively, which justified the design of the ECM. In addition, an average improvement of 1.55% and 1.81% in segmentation accuracy were achieved by using SBSS-MS and SBSS-SS, respectively.

To visually validate the effectiveness of the proposed SBSS framework, a comparison of the segmentation results generated by MS and SBSS-MS is shown in Fig. 4. It can be observed that the segmentation results obtained using SBSS-MS had less visually fragmented areas compared to the MS (e.g., the areas within the red box in the example images of UAVid, LoveDA and Potsdam). In the meantime, the segmentation example from Cityscapes showed that SBSS-MS was also able to segment objects that were completely missed by MS.

### E. The complexity and the speed of the SBSS framework

For a comprehensive comparison of the efficiency of SBSS with the other methods, experiments were conducted on the Cityscapes validation set, the results of which are recorded in Table IX. All models in the experiments were implemented using the PyTorch framework. The training time was measured with four NVIDIA GTX 3090 GPU. The inference speed was measured in terms of the number of tasks (original images rather than single patches) per second, and calculated as the average value of 500 tests on a single NVIDIA GTX 3090 GPU.

For the multi-scale test level comparisons, the SBSS-MS achieved both the highest speed (0.85 Task/s) and the highest accuracy (mIoU of 83.05%).

In the single scale test level comparisons, the input patch size of the other methods was set to four times the size of the SBSS-SS. The rationale for this setting is as follows. It was noticed that the segmentation accuracy decreased when using smaller patches, as shown in Table VII. Meanwhile, Table VIII shows that SBSS-SS improved the segmentation accuracy compared to SS. In addition, the memory footprint was proportional to the patch size. For example, with the same network, a memory footprint with a patch size of 512 x 256 is a quarter of the one that uses a patch size of 1024 x 512. Therefore, it is meaningful to test whether SBSS-SS using a smaller patch size can achieve similar accuracy to other methods using a larger patch size. The results in Table IX show that this can be achieved with SBSS-SS, which achieves an mIoU (78.52%) that is merely (0.32%) lower than the highest one (78.84%).

## F. Quantitative results on Cityscapes, UAVid, LoveDA and Potsdam test sets

To further confirm the effectiveness of the proposed SBSS, experimental comparison between the SBSS and other state-of-the-art methods was conducted on the test set of each dataset considered. The input patch sizes used for the different datasets were the same as those listed in Table VIII. Apart from the Potsdam dataset that was evaluated offline, the segmentation results were submitted to the online servers dedicated for other dataset for evaluation, and the performance results are summarised in Tables X-XIII.

Because of the strict limitations on the test frequency in the Cityscapes test server, we submitted only the segmentation results from our own methods (SBSS-MS and SBSS-SS). The performance results of the other methods on the Cityscapes dataset were taken directly from their original publications. It is worth mentioning that since these methods were obtained with different training sets, backbones, training and test setups, it is probably not very rigours to simply compare the results in Table X. Nevertheless, those results show that SBSS-MS achieved the highest segmentation accuracy, which confirms the effectiveness of our method.

The UAVid and the LoveDA datasets are less restricted in terms of the test frequency in the test servers. The Potsdam dataset can be tested offline. As such, for fairer comparisons, the segmentation results from the applications of all the methods (including ours and the others) to these three datasets were obtained by ourselves using the same settings for training and test. The training and test settings were the same as those used in Sections III-A-2 and III-D, respectively. The proposed SBSS-MS achieved the highest segmentation accuracy at the multi-scale test level on all three datasets. At the same time, SBSS-SS achieved comparable segmentation accuracy to the other methods using a smaller patch size (i.e., smaller memory footprint) at the single scale test level on all three datasets.\

TABLE X
QUANTITATIVE COMPARISON RESULTS ON THE CITYSCAPES TEST SET. THE INPUT PATCH SIZES USED IN SBSS-MS AND SBSS-SS ARE 1024×512 AND 512×256 RESPECTIVELY

| Method | Backbone | Trained on | Test method | mIoU (%) |
|---|---|---|---|---|
| PSPNet | ResNet101 | Train | MS | 78.4 |
| BiSeNetV1 | ResNet101 | Train & Val | SS | 78.9 |
| PSANet | ResNet101 | Train & Val | MS | 80.1 |
| DenseASPP | DenseNet201 | Train & Val | MS | 80.6 |
| SETR | ViT-L | Train & Val | MS | 81.1 |
| Segmenter | ViT-L | Train & Val | MS | 81.3 |
| DANet | ResNet101 | Train & Val | MS | 81.5 |
| HRNet | HRNet-w48 | Train & Val | MS | 81.6 |
| EANet | ResNet101 | Train & Val | MS | 81.7 |
| OCR | ResNet101 | Train & Val | MS | 81.8 |
| DNL | ResNet101 | Train & Val | MS | 82.0 |
| SegFormer | MiT-B5 | Train & Val | MS | 82.2 |
| SBSS-SS | ConvNeXt-T | Train & Val | ECS-SS | 80.3 |
| SBSS-MS | ConvNeXt-T | Train & Val | ECS-MS | 82.6 |

TABLE IX
COMPARISON OF THE EFFICIENCY OF THE SBSS FRAMEWORK WITH OTHER METHODS ON CITYSCAPES VALIDATION SET

| Method | Backbone | Test method | Patch size | mIoU (%) | Parameters (M) | Flops per patch (G) | Flops per task (G) | Training Time (h) | Task/s |
|---|---|---|---|---|---|---|---|---|---|
| FCN | HRNet-w18 | MS | 1024×512 | 79.47 | 48.98 | 356.91 | 13919.49 | 9.88 | 0.44 |
| BiSeNetV1 | ResNet50 | MS | 1024×512 | 79.31 | 59.24 | 197.91 | 7718.49 | 10.73 | 0.81 |
| PSPNet | ResNet50 | MS | 1024×512 | 80.55 | 48.98 | 356.91 | 13919.49 | 15.79 | 0.54 |
| DeepLabV3+ | ResNet50 | MS | 1024×512 | 81.18 | 43.59 | 352.72 | 13756.08 | 16.32 | 0.49 |
| DANet | ResNet50 | MS | 1024×512 | 81.09 | 49.85 | 398.30 | 15533.70 | 15.18 | 0.48 |
| GCNet | ResNet50 | MS | 1024×512 | 80.40 | 49.63 | 395.46 | 15422.94 | 19.35 | 0.53 |
| DNLNet | ResNet50 | MS | 1024×512 | 80.73 | 50.02 | 399.76 | 15590.64 | 14.53 | 0.47 |
| UperNet | ResNet50 | MS | 1024×512 | 80.17 | 66.42 | 473.65 | 18472.35 | 14.17 | 0.50 |
| UperNet | Swin-T | MS | 1024×512 | 79.99 | 59.84 | 469.04 | 18292.56 | 14.25 | 0.48 |
| UperNet | ConvNeXt-T | MS | 1024×512 | 81.32 | 60.14 | 467.15 | 18218.85 | 18.25 | 0.53 |
| SBSS-MS | ConvNeXt-T | ECS-MS | 1024×512 | 83.05 | 60.99 | 490.51 | 11281.76 | 22.18 | 0.85 |
| FCN | HRNet-w18 | SS | 1024×512 | 75.73 | 48.98 | 356.91 | 1427.64 | 9.88 | 4.31 |
| BiSeNetV1 | ResNet50 | SS | 1024×512 | 75.06 | 59.24 | 197.91 | 791.64 | 10.73 | 7.91 |
| PSPNet | ResNet50 | SS | 1024×512 | 77.90 | 48.98 | 356.91 | 1427.64 | 15.79 | 5.25 |
| DeepLabV3+ | ResNet50 | SS | 1024×512 | 78.66 | 43.59 | 352.72 | 1410.88 | 16.32 | 4.74 |
| DANet | ResNet50 | SS | 1024×512 | 78.64 | 49.85 | 398.30 | 1593.20 | 15.18 | 4.72 |
| GCNet | ResNet50 | SS | 1024×512 | 77.68 | 49.63 | 395.46 | 1581.84 | 19.35 | 5.14 |
| DNLNet | ResNet50 | SS | 1024×512 | 78.31 | 50.02 | 399.76 | 1599.04 | 14.53 | 4.55 |
| UperNet | ResNet50 | SS | 1024×512 | 77.65 | 66.42 | 473.65 | 1894.60 | 14.17 | 4.89 |
| UperNet | Swin-T | SS | 1024×512 | 77.47 | 59.84 | 469.04 | 1876.16 | 14.25 | 4.63 |
| UperNet | ConvNeXt-T | SS | 1024×512 | 78.84 | 60.14 | 467.15 | 1868.60 | 18.25 | 5.12 |
| SBSS-SS | ConvNeXt-T | ECS-SS | 512×256 | 78.52 | 60.78 | 123.79 | 1485.54 | 21.53 | 6.50 |

TABLE XI
QUANTITATIVE COMPARISON RESULTS ON THE UAVID TEST SET (%)

| Method | Backbone | Test method | Patch size | mIoU | Building | Static Car | Tree | Moving Car | Clutter | Road | Human | Vegetation |
|--------|----------|-------------|------------|------|----------|------------|------|------------|---------|------|-------|------------|
| FCN | HRNet-w18 | MS | 1024×540 | 71.11 | 89.85 | 69.08 | 81.86 | 78.03 | 71.80 | 83.89 | 28.08 | 66.30 |
| BiSeNetV1 | ResNet50 | MS | 1024×540 | 69.03 | 88.57 | 65.03 | 81.47 | 73.18 | 69.64 | 82.20 | 26.55 | 65.57 |
| PSPNet | ResNet50 | MS | 1024×540 | 69.76 | 88.96 | 63.84 | 81.34 | 75.84 | 70.61 | 82.81 | 29.37 | 65.27 |
| DeepLabV3+ | ResNet50 | MS | 1024×540 | 71.06 | 89.44 | 71.10 | 81.65 | 77.23 | 71.01 | 82.81 | 29.43 | 65.84 |
| DANet | ResNet50 | MS | 1024×540 | 70.22 | 89.36 | 66.43 | 81.46 | 75.53 | 70.96 | 82.83 | 29.60 | 65.59 |
| GCNet | ResNet50 | MS | 1024×540 | 69.62 | 89.24 | 63.80 | 81.25 | 74.87 | 70.69 | 82.69 | 29.34 | 65.06 |
| DNLNet | ResNet50 | MS | 1024×540 | 69.67 | 88.89 | 63.74 | 81.58 | 75.58 | 70.42 | 82.74 | 28.84 | 65.58 |
| UperNet | ResNet50 | MS | 1024×540 | 70.87 | 89.35 | 68.66 | 81.68 | 77.30 | 71.00 | 83.10 | 30.20 | 65.67 |
| UperNet | Swin-tiny | MS | 1024×540 | 70.43 | 89.17 | 65.11 | 81.33 | 78.16 | 70.43 | 82.71 | 29.66 | 65.63 |
| UperNet | ConvNeXt-T | MS | 1024×540 | 71.22 | 89.81 | 69.57 | 81.78 | 77.92 | 71.14 | 82.57 | 30.83 | 66.11 |
| SBSS-MS | ConvNeXt-T | ECS-MS | 1024×540 | 72.99 | 91.05 | 76.00 | 82.50 | 78.20 | 73.06 | 83.57 | 31.95 | 67.59 |
| FCN | HRNet-w18 | SS | 1024×540 | 69.34 | 88.90 | 66.29 | 80.45 | 75.94 | 69.62 | 81.91 | 27.82 | 63.78 |
| BiSeNetV1 | ResNet50 | SS | 1024×540 | 67.31 | 87.32 | 63.11 | 80.06 | 70.26 | 67.47 | 80.40 | 26.62 | 63.25 |
| PSPNet | ResNet50 | SS | 1024×540 | 68.23 | 88.09 | 61.47 | 80.13 | 73.98 | 69.04 | 81.19 | 28.65 | 63.25 |
| DeepLabV3+ | ResNet50 | SS | 1024×540 | 69.57 | 88.56 | 68.21 | 80.43 | 75.98 | 69.27 | 81.22 | 29.17 | 63.69 |
| DANet | ResNet50 | SS | 1024×540 | 68.61 | 88.48 | 63.27 | 80.31 | 73.38 | 69.47 | 81.56 | 28.70 | 63.75 |
| GCNet | ResNet50 | SS | 1024×540 | 67.95 | 88.27 | 61.03 | 79.84 | 72.52 | 69.00 | 81.19 | 29.07 | 62.71 |
| DNLNet | ResNet50 | SS | 1024×540 | 68.14 | 87.92 | 60.57 | 80.35 | 73.50 | 68.80 | 81.47 | 28.89 | 63.63 |
| UperNet | ResNet50 | SS | 1024×540 | 69.24 | 88.35 | 66.18 | 80.31 | 75.60 | 69.23 | 81.53 | 29.67 | 63.03 |
| UperNet | Swin-tiny | SS | 1024×540 | 69.13 | 88.44 | 64.04 | 80.33 | 76.14 | 69.14 | 81.74 | 29.33 | 63.89 |
| UperNet | ConvNeXt-T | SS | 1024×540 | 70.05 | 89.22 | 67.64 | 80.91 | 76.07 | 69.84 | 81.34 | 30.59 | 64.77 |
| SBSS-MS | ConvNeXt-T | ECS-SS | 512×270 | 70.00 | 88.19 | 70.50 | 81.37 | 76.08 | 68.80 | 82.31 | 27.50 | 65.25 |

TABLE XII
QUANTITATIVE COMPARISON RESULTS ON THE LOVEDA TEST SET (%)

| Method | Backbone | Test method | Patch size | mIoU | Background | Building | Road | Water | Barren | Forest | Agricultural |
|--------|----------|-------------|------------|------|------------|----------|------|-------|--------|--------|--------------|
| FCN | HRNet-w18 | MS | 512×512 | 52.74 | 45.33 | 59.60 | 56.26 | 80.62 | 17.81 | 48.92 | 60.64 |
| BiSeNetV1 | ResNet50 | MS | 512×512 | 50.46 | 44.73 | 55.36 | 55.52 | 77.85 | 14.07 | 45.80 | 59.87 |
| PSPNet | ResNet50 | MS | 512×512 | 52.43 | 45.42 | 57.50 | 58.96 | 79.24 | 17.98 | 48.66 | 59.21 |
| DeepLabV3+ | ResNet50 | MS | 512×512 | 52.55 | 44.99 | 56.88 | 59.35 | 79.19 | 18.41 | 48.83 | 60.19 |
| DANet | ResNet50 | MS | 512×512 | 50.92 | 44.05 | 54.15 | 54.97 | 77.62 | 19.33 | 47.12 | 59.17 |
| GCNet | ResNet50 | MS | 512×512 | 52.76 | 45.80 | 58.30 | 57.94 | 79.55 | 18.47 | 48.50 | 60.76 |
| DNLNet | ResNet50 | MS | 512×512 | 52.59 | 45.33 | 57.13 | 57.59 | 79.60 | 19.01 | 48.23 | 61.27 |
| UperNet | ResNet50 | MS | 512×512 | 52.30 | 45.44 | 57.32 | 59.17 | 79.16 | 18.00 | 47.66 | 59.38 |
| UperNet | Swin-tiny | MS | 512×512 | 53.22 | 46.29 | 58.66 | 58.86 | 80.91 | 17.88 | 47.88 | 62.03 |
| UperNet | ConvNeXt-T | MS | 512×512 | 53.57 | 46.51 | 60.26 | 59.95 | 80.53 | 17.12 | 48.14 | 62.50 |
| SBSS-MS | ConvNeXt-T | ECS-MS | 512×512 | 54.50 | 46.31 | 62.35 | 58.66 | 82.06 | 19.59 | 49.48 | 63.07 |
| FCN | HRNet-w18 | SS | 512×512 | 51.08 | 43.78 | 57.56 | 54.33 | 78.59 | 16.95 | 47.13 | 59.24 |
| BiSeNetV1 | ResNet50 | SS | 512×512 | 48.67 | 42.50 | 53.38 | 53.62 | 76.93 | 14.03 | 42.79 | 57.42 |
| PSPNet | ResNet50 | SS | 512×512 | 50.63 | 44.15 | 54.72 | 56.54 | 76.81 | 17.49 | 47.13 | 57.54 |
| DeepLabV3+ | ResNet50 | SS | 512×512 | 50.54 | 43.35 | 54.40 | 56.96 | 76.61 | 17.65 | 47.08 | 57.69 |
| DANet | ResNet50 | SS | 512×512 | 49.18 | 42.18 | 42.18 | 58.29 | 56.23 | 20.36 | 48.79 | 61.78 |
| GCNet | ResNet50 | SS | 512×512 | 50.82 | 44.47 | 55.55 | 55.63 | 77.35 | 17.54 | 46.64 | 58.56 |
| DNLNet | ResNet50 | SS | 512×512 | 50.56 | 43.85 | 54.22 | 54.88 | 77.04 | 17.51 | 46.68 | 59.76 |
| UperNet | ResNet50 | SS | 512×512 | 50.27 | 43.75 | 54.81 | 56.58 | 76.93 | 17.20 | 45.75 | 56.87 |
| UperNet | Swin-tiny | SS | 512×512 | 51.63 | 44.85 | 55.96 | 56.54 | 80.06 | 17.87 | 45.58 | 60.62 |
| UperNet | ConvNeXt-T | SS | 512×512 | 52.19 | 44.89 | 62.05 | 59.16 | 79.75 | 16.74 | 47.22 | 55.55 |
| SBSS-SS | ConvNeXt-T | ECS-SS | 256×256 | 52.15 | 44.69 | 58.88 | 58.32 | 79.14 | 16.52 | 46.68 | 60.82 |

### G. Qualitative Analysis of the Segmentation Results

As introduced in Section I, the objective of the proposed SBSS is to reduce the error associated with the resizing scales. While the efficiency and the effectiveness of SBSS were demonstrated in Section III-D to Section III-F, it is informative to appreciate what kind of errors associated with the resizing scales were corrected by SBSS. The class building was chosen for analysis because the segmentation of buildings is crucial for many applications and this class happens to be present in all four datasets used.

The top four plots of Fig. 5 show the segmentation accuracy (IoU) of buildings for the four datasets when they were tested with SS at different resizing scales. It was observed that the segmentation accuracies of buildings in the Cityscapes, UAVid and Potsdam datasets were generally higher when smaller resizing scales were used. This is consistent with the fact that these datasets were collected in urban scenes, where buildings are expected to be the relatively large objects. However, as the LoveDA dataset includes many images of rural scenes where buildings are comparatively smaller objects than forests and water bodies, a larger resizing scale was favourable.

In addition, the segmentation accuracies of the MS and SBSS-MS were tested as follows. The scale started with the smallest one (i.e., 0.5), followed by a stepwise increase (with an increment of 0.25 each time) until the largest scale of 1.75 was reached. In each test, all the scales that are equal to or smaller than a particular scale were used. For example, for the scale 1, the following scales 0.5, 0.75 and 1 were used. The results are shown in the bottom four plots of Fig. 5, suggesting that the accuracy of MS did not always increase when more and larger scales were used. For example, the accuracy of MS on the UAVid and Potsdam datasets decreased after using scales larger than 0.75 and 1.5, respectively. To investigate the causes of this phenomenon, the segmentation results of a UAVid image using SS at six resizing scales and those for MS and SBSS-MS using the scales from 0.5 to 1.75 are plotted in Fig. 6.

TABLE XIII
QUANTITATIVE COMPARISON RESULTS ON THE POTSDAM TEST SET (%)

| Method | Backbone | Test method | Patch size | mIoU | Impervious surface | Building | Low vegetation | Tree | Car |
|---|---|---|---|---|---|---|---|---|---|
| FCN | HRNet-w18 | MS | 750×750 | 86.79 | 88.10 | 94.06 | 78.85 | 80.78 | 92.17 |
| BiSeNetV1 | ResNet50 | MS | 750×750 | 86.37 | 87.74 | 93.53 | 78.43 | 80.24 | 91.90 |
| PSPNet | ResNet50 | MS | 750×750 | 85.81 | 87.26 | 93.37 | 77.14 | 79.59 | 91.68 |
| DeepLabV3+ | ResNet50 | MS | 750×750 | 86.85 | 88.20 | 94.16 | 78.37 | 80.79 | 92.75 |
| DANet | ResNet50 | MS | 750×750 | 86.85 | 87.87 | 93.76 | 78.53 | 80.99 | 93.09 |
| GCNet | ResNet50 | MS | 750×750 | 86.91 | 88.35 | 93.83 | 78.52 | 80.95 | 92.90 |
| DNLNet | ResNet50 | MS | 750×750 | 86.76 | 88.18 | 93.75 | 78.08 | 80.89 | 92.91 |
| UperNet | ResNet50 | MS | 750×750 | 86.84 | 88.27 | 93.87 | 78.52 | 80.96 | 92.56 |
| UperNet | Swin-tiny | MS | 750×750 | 87.01 | 88.52 | 94.31 | 79.06 | 81.16 | 92.01 |
| UperNet | ConvNeXt-T | MS | 750×750 | 87.34 | 88.82 | 94.58 | 79.27 | 81.60 | 92.44 |
| SBSS-MS | ConvNeXt-T | ECS-MS | 750×750 | 87.68 | 88.95 | 94.88 | 79.42 | 81.82 | 93.34 |
| FCN | HRNet-w18 | SS | 750×750 | 85.49 | 87.30 | 92.97 | 77.32 | 79.45 | 90.43 |
| BiSeNetV1 | ResNet50 | SS | 750×750 | 84.66 | 86.23 | 92.34 | 76.54 | 78.11 | 90.06 |
| PSPNet | ResNet50 | SS | 750×750 | 85.81 | 87.26 | 93.37 | 77.14 | 79.59 | 91.68 |
| DeepLabV3+ | ResNet50 | SS | 750×750 | 85.68 | 87.32 | 93.43 | 76.83 | 79.33 | 91.51 |
| DANet | ResNet50 | SS | 750×750 | 86.01 | 87.49 | 93.48 | 77.16 | 79.86 | 92.05 |
| GCNet | ResNet50 | SS | 750×750 | 85.82 | 87.47 | 93.24 | 77.04 | 79.67 | 91.70 |
| DNLNet | ResNet50 | SS | 750×750 | 85.61 | 87.21 | 93.10 | 76.35 | 79.54 | 91.84 |
| UperNet | ResNet50 | SS | 750×750 | 85.62 | 87.28 | 92.90 | 77.09 | 79.64 | 91.17 |
| UperNet | Swin-tiny | SS | 750×750 | 86.07 | 87.88 | 93.75 | 77.92 | 79.99 | 90.81 |
| UperNet | ConvNeXt-T | SS | 750×750 | 86.41 | 88.07 | 94.04 | 78.15 | 80.52 | 91.27 |
| SBSS-SS | ConvNeXt-T | ECS-SS | 375×375 | 86.35 | 87.93 | 93.82 | 78.06 | 80.55 | 91.40 |

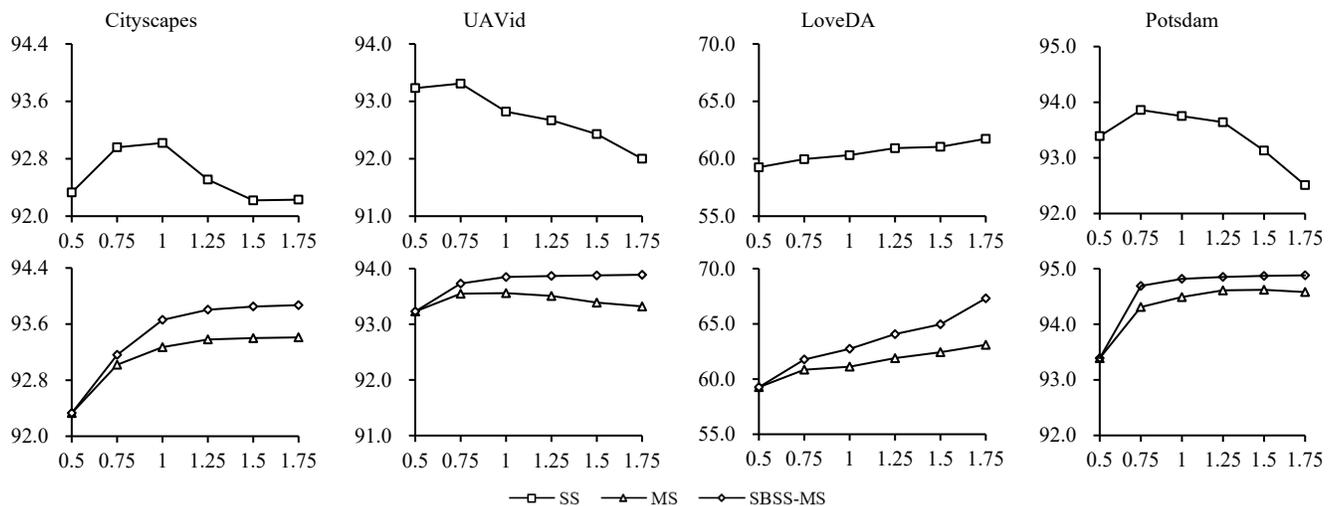

**Fig. 5.** Segmentation accuracy (IoU %) of buildings using different resizing scales. The horizontal coordinates (of the top 4 plots) for SS refer to the resizing scale used. The horizontal coordinates (of the bottom 4 plots) for MS and SBSS represent the utilisation of all the scales that are equal to and smaller than the current scale (e.g., the coordinate value 1 represents the case where the scales 0.5, 0.75 and 1 were all used).

As shown in Fig.6, when the scale increased, SS yielded more fragmented and erroneous results for the large building on the right. The likely reasons for this are presented in the following. First, the use of a large resizing scale introduced additional difficulties for the global context information modelling. Second, the complex building facade in Fig. 6 made correct segmentation more dependent on modelling global information rather than local one. Since MS never learns at which scale the segmentation results are more reliable for the building, the final results obtained using the average voting inevitably inherit some of the errors in the segmentation results at relatively large scales. In contrast, SBSS-MS preserved the correct segmentation results for buildings in a more intelligent way.

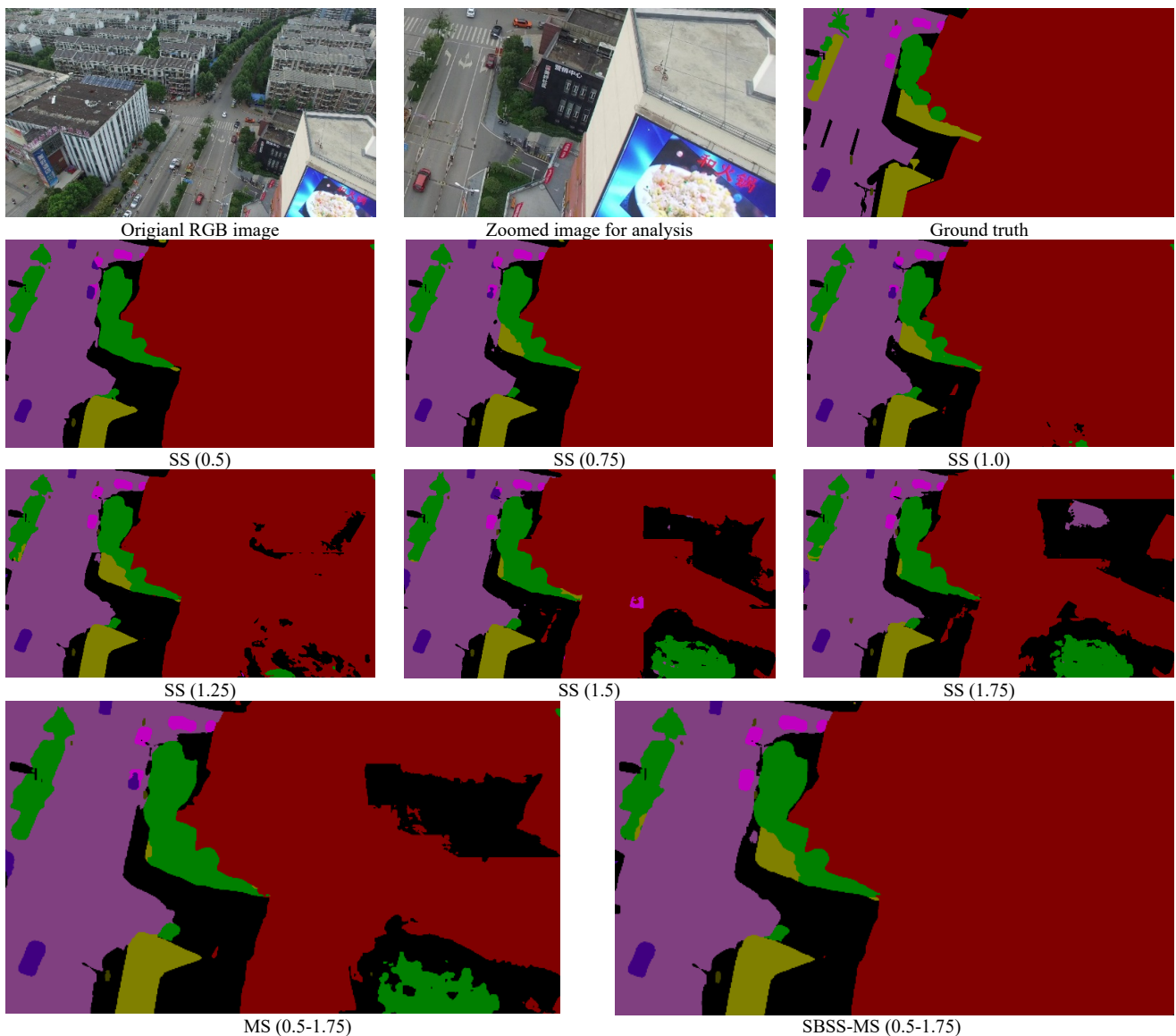

**Fig. 6.** Visual comparisons between SS, MS and SBSS-MS on the UAVid validation set. The numbers in brackets represent the resizing scales used.

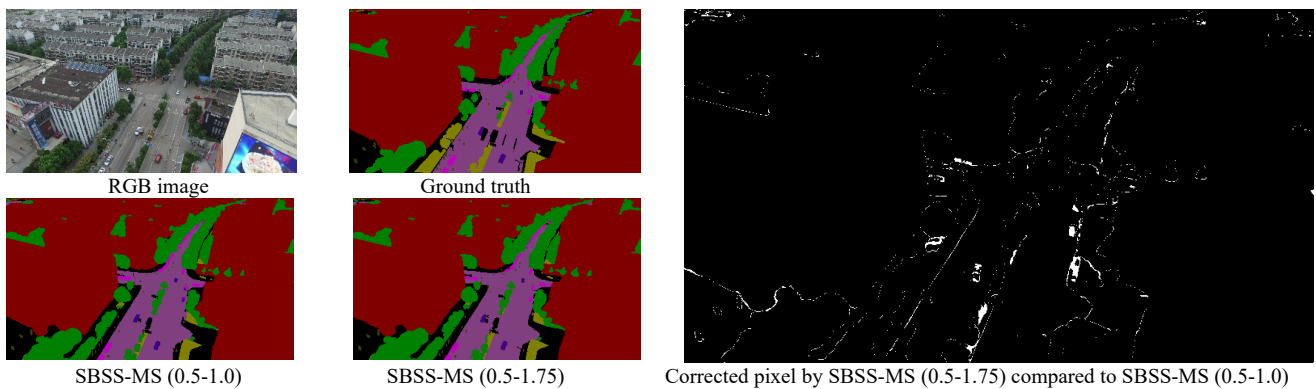

**Fig. 7.** Visual comparisons between SBSS-MS using different set of resizing scales on the UAVid validation set.

Apart from this type of erroneous segmentation of a large area of the building, there is another common type of error that occurs at the edges of the building. To verify whether our SBSS-MS can handle this type of error, a comparison was made between results generated by SBSS-MS using two different sets of resizing scales (i.e., 0.5-1.0 and 0.5-1.75), and the results are shown in Fig. 7. It shows that using the SS results at three larger scales (i.e., 1.25, 1.5 and 1.75), SBSS-MS improved the segmentation accuracy at the building edges.

TABLE XIV
QUANTITATIVE COMPARISON FOR SBSS-MS USING DIFFERENT INPUT METHODS ON THE LOVEDA TEST SET (%)

| Method | Backbone | Patch size | Scales used | mIoU | Background | Building | Road | Water | Barren | Forest | Agricultural |
|---|---|---|---|---|---|---|---|---|---|---|---|
| SBSS-MS | ConvNeXt-T | 512×512 | 0.5-1.5 | 54.50 | 46.31 | 62.35 | 58.66 | 82.06 | 19.59 | 49.48 | 63.07 |
| SBSS-MS | ConvNeXt-T | Entire image | 0.5-1.5 | 54.71 | 46.92 | 62.20 | 58.17 | 82.24 | 19.78 | 49.41 | 64.22 |
| SBSS-MS | ConvNeXt-T | Entire image | 0.5-2.0 | 55.59 | 48.30 | 62.85 | 58.22 | 83.06 | 21.02 | 49.76 | 65.93 |

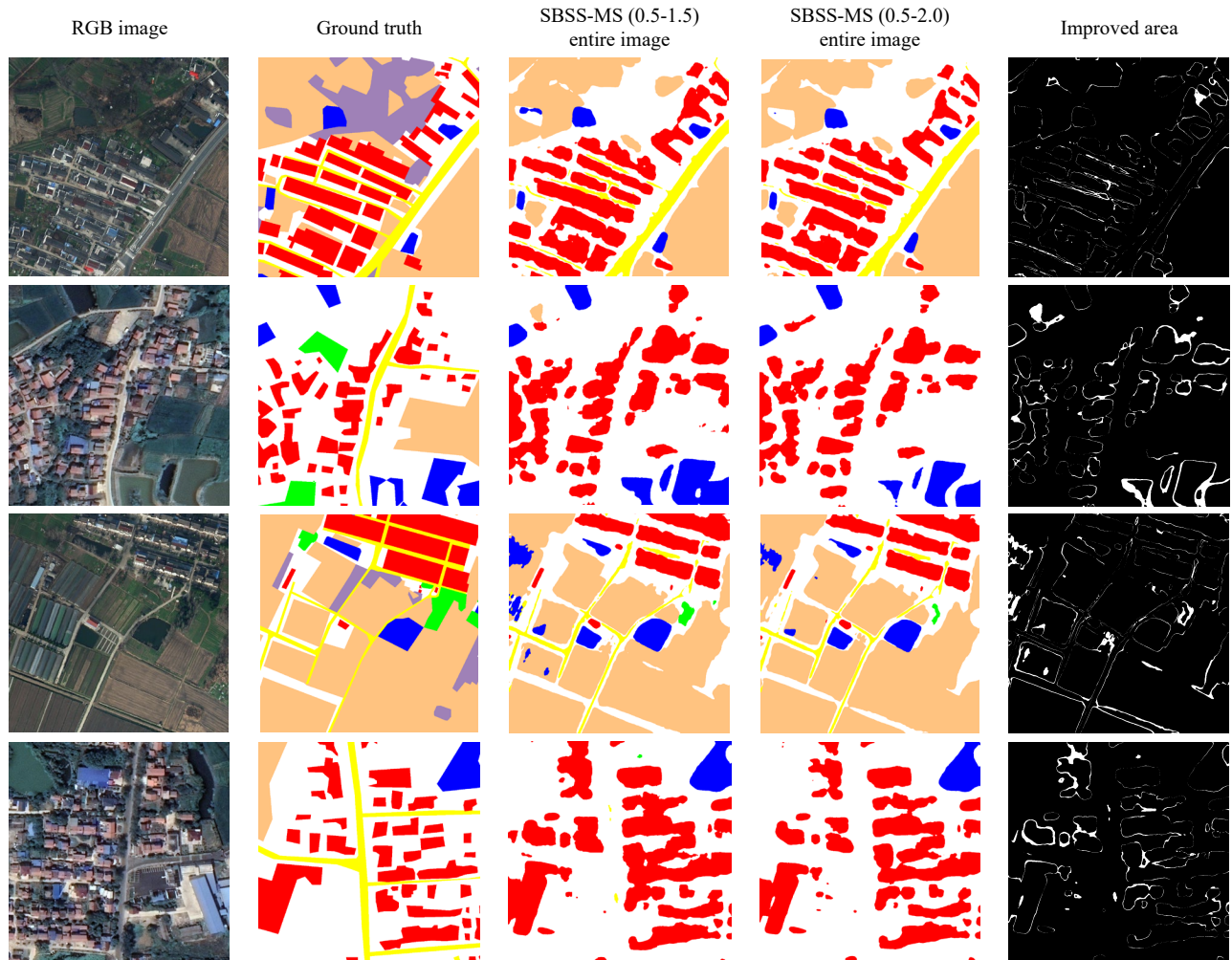

**Fig. 8.** Qualitative comparisons between SBSS-MS with different setting on the LOVEDA validation set.

Also, it is worth mentioning that although building was used as the example class for the demonstration, the aforementioned two kinds of improvements are expected to be applicable to the other classes. For example, SBSS-MS corrected the car that was mis-segmented in MS (the top left part in Fig. 6). The segmentation improvement at the object edges in Fig. 7 was also valid for other classes such as tree and road.

*H. Exploring the potential for higher accuracy*

The computational complexity of SBSS-MS has been strictly limited in the previous sections (i.e., Section III-D to Section III-F) to facilitate a fair comparison with the other methods. However, it is of interest to discover the accuracy that can be achieved by SBSS-MS when adequate computational resources are available.

The results in Table VII show that the segmentation accuracy decreased considerably when smaller patch sizes were used.

Therefore, our study tested the case of directly performing the segmentation on the entire image in the first place. As the memory of the GPU used is limited, only the LoveDA dataset, which has a relatively small original image size, was tested. In addition, on the basis of using the whole image for segmentation, the case of using more resizing scales (i.e., 0.5-2.0) was tested.

The quantitative results of the test are shown in Table XIV, suggesting that using the whole image for segmentation did improve the segmentation accuracy slightly (0.21% in mIoU), but using more scales was a more effective way (1.09% in mIoU) in comparison. The segmentation results for the second and third settings in Table XIV show that by using larger resizing scales SBSS-MS improved the segmentation accuracy for all classes. Meanwhile, Fig. 8 suggest that the improvement was mainly at the edges of the objects.

## IV. Future work

In the current study, the final segmentation result is obtained by fusing the segmentation results at a predefined set of resizing. However, the results in Tables VI indicate that the optimal scale for each class varies from case to case. Therefore, the performance of SBSS could be further improved by developing algorithms that can adaptively select a set of scales for analyses. Similarly, it can be speculated that such adaptive algorithms would also be useful for choosing how many patches to analyse at each scale. In the course of this study, it was noticed that the error distribution has a pattern not only in the dimension of resizing scales, but also in the spatial dimension. For example, errors are more likely to occur at the edge of a patch. This characteristic could be taken into consideration in future study.

In addition to an improvement of the SBSS framework itself, future research could consider combining the SBSS framework with other existing methods such as edge-aware segmentation [42]–[45] or object-based segmentation [29], [46]. When visually inspecting the differences between ground truth, MS and SBSS-MS segmentation results (e.g., Fig. 4 and Fig. 8), it was found that although SBSS-MS could obtain more accurate segmentation results than MS, there are still cases where an object is segmented into pieces. Therefore, integration with those studies (e.g., [29], [42]–[46]) that specifically address this issue may further improve segmentation accuracy.

## V. Conclusion

This study experimentally demonstrated that different classes in images have their preferred resizing scales for semantic segmentations. On this basis, the SBSS framework was proposed, which uses a learnable ECM to fuse segmentation results that are more likely to be correct at each resizing scale, and an ECS to control the computational complexity. Extensive experiments were conducted on the four benchmark datasets considered, i.e., Cityscapes, UAVid, LoveDA and Potsdam datasets. The results show that SBSS achieved promising performances in the various scenarios considered. Specifically, SBSS-MS achieved a higher segmentation accuracy with less Flops, faster speed, and similar memory footprint compared to MS. Meanwhile, SBSS-SS achieved a similar segmentation accuracy with a quarter of the memory footprint, similar Flops and speed compared to SS. In the future, more sophisticated ECS and ECM can be proposed to further improve the performance of SBSS or to adapt it to specific application requirements.